\documentclass[runningheads]{llncs}

\usepackage{eccv}

\usepackage{eccvabbrv}
\usepackage[toc,title,page]{appendix}
\usepackage{multirow}
\usepackage{graphicx}
\usepackage{booktabs}

\usepackage[accsupp]{axessibility}  

\usepackage{hyperref}

\usepackage{orcidlink}

\begin{document}

\title{Rethinking and Improving Visual Prompt Selection for In-Context Learning Segmentation} 

\titlerunning{Visual Prompt Selection for In-Context Learning Segmentation}

\author{Wei Suo\inst{1,3,\footnotemark[1]}\orcidlink{0000-0002-8283-8637} \and
Lanqing Lai\inst{1,\footnotemark[1]} \and
Mengyang Sun\inst{2,3}\orcidlink{0000-0003-0638-3295} \and
Hanwang Zhang\inst{3}\orcidlink{0000-0001-7374-8739} \and
Peng Wang\inst{1,\footnotemark[4]} \and 
Yanning Zhang\inst{1}\orcidlink{0000-0002-2977-8057}
}

\authorrunning{W. Suo et al.}

\institute{School of Computer Science and Ningbo Institute, Northwestern Polytechnical University, Xi'an, China \and
School of Cybersecurity, Northwestern Polytechnical University, Xi'an, China 
\and
School of Computer Science, Nanyang Technological University, Singapore
\email{\{suowei1994,lailanqing,sunmenmian\}@mail.nwpu.edu.cn}\\
}

\maketitle

\renewcommand{\thefootnote}{\fnsymbol{footnote}}
\footnotetext[4]{Corresponding author}
\footnotetext[1]{These authors contributed equally to this work.}
\begin{abstract}
  As a fundamental and extensively studied task in computer vision, image segmentation aims to locate and identify different semantic concepts at the pixel level. Recently, inspired by In-Context Learning (ICL), several generalist segmentation frameworks have been proposed, providing a promising paradigm for segmenting specific objects. However, existing works mostly ignore the value of visual prompts or simply apply similarity sorting to select contextual examples. In this paper, we focus on rethinking and improving the example selection strategy. 
  By comprehensive comparisons, we first demonstrate that ICL-based segmentation models are sensitive to different contexts. Furthermore, empirical evidence indicates that the diversity of contextual prompts plays a crucial role in guiding segmentation. Based on the above insights, we propose a new stepwise context search method. Different from previous works, we construct a small yet rich candidate pool and adaptively search the well-matched contexts. More importantly, this method effectively reduces the annotation cost by compacting the search space. Extensive experiments show that our method is an effective strategy for selecting examples and enhancing segmentation performance. \href{https://github.com/LanqingL/SCS}{https://github.com/LanqingL/SCS}
  \keywords{{In-Context Learning  \and Visual Prompt \and Image Segmentation}
  }
\end{abstract}

\section{Introduction}
\label{sec:intro}

Image segmentation is a fundamental task in computer vision, aiming to locate and identify different semantic concepts at the pixel level. 
It is essential in multiple applications such as 
autonomous driving~\cite{wang2020deep,zendel2022unifying}, video surveillance~\cite{vasudevan2020semantic,li2020reducto,sun2024adaptive,suo2022simple} and complex reasoning~\cite{shi2018key,luo2020multi,suo2022rethinking}.
To achieve precise segmentation, researchers have proposed a variety of segmentation models~\cite{ronneberger2015u,long2015fully,he2017mask,badrinarayanan2017segnet}. Although these methods have made considerable progress, they require training the specialist model for each segmentation task.
Recently, inspired by In-Context Learning (ICL)~\cite{wei2022chain,min2022rethinking,rubin2021learning}, several generalist segmentation frameworks have been proposed and achieved impressive performance such as MAE-VQGAN~\cite{bar2022visual} and SegGPT~\cite{wang2023SegGPT}. 
Different from the traditional frameworks, ICL-based segmentation methods establish a promising paradigm for localizing vision concepts by feeding one or a few examples as references during inference. Meanwhile, the methodology offers a general solution for various real-world scenarios~\cite{wang2023images}. 

\begin{figure*}[t]
\centering
    \includegraphics[width=0.75\textwidth]{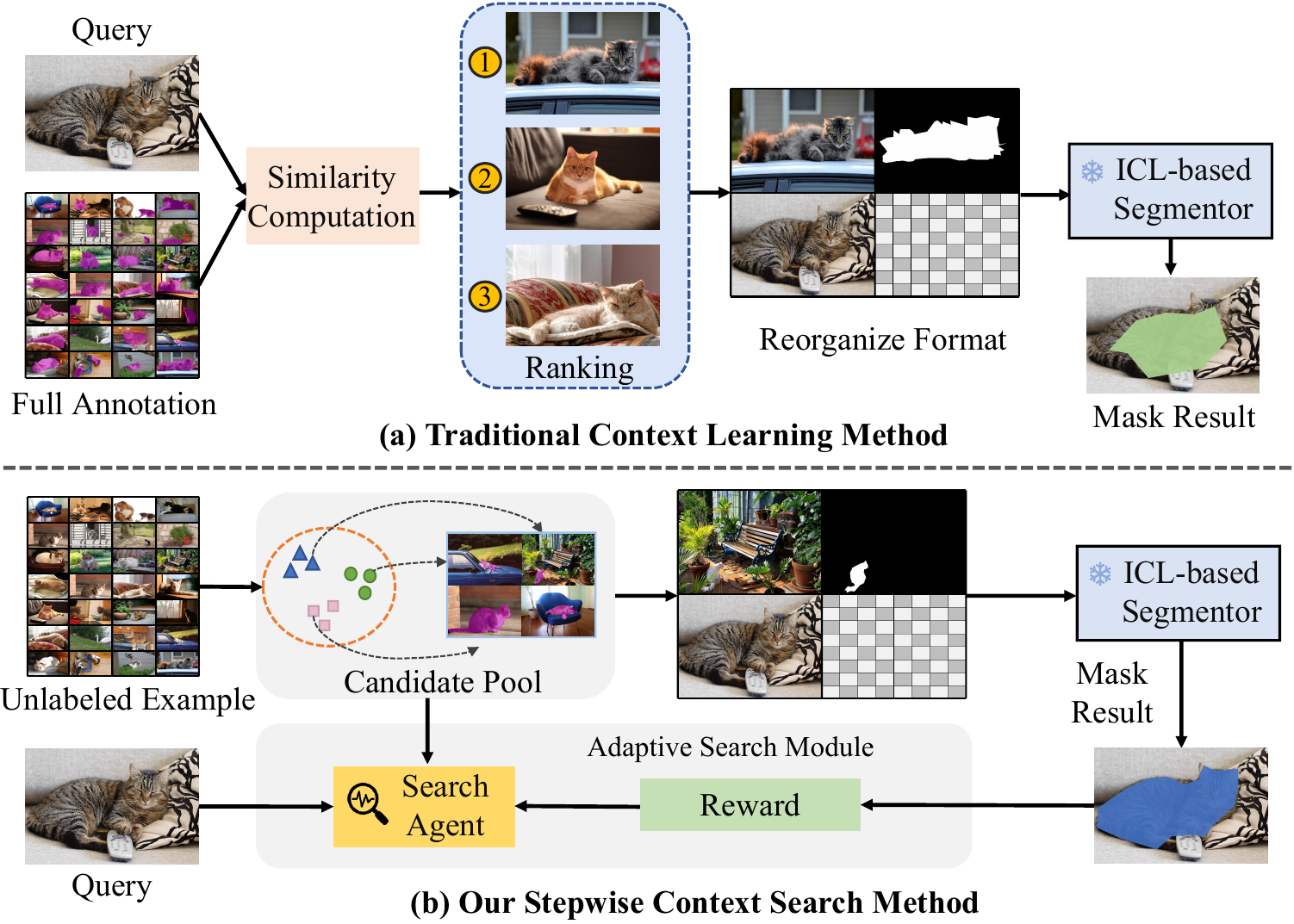}
    \caption{Comparsion of traditional and our method. (a) Existing works rely on dense annotation to build the exampling space. Then, they use the similarity sorting manner to select examples for given queries.
    (b) Instead, our method significantly alleviates the costs of annotation by searching the typical examples. Moreover, a novel adaptive search module is designed to further select well-matched contexts.}
    \label{fig:motivation}
\end{figure*}

ICL learning is a relatively new research line in computer vision, while the fundamental ideas can be traced back to the Natural Language Processing (NLP) field~\cite{bar2022visual,wang2023images}. With the development of Large Language Models (LLM), numerous studies have already proved that selecting a suitable context is crucial for following human instruction and improving performance~\cite{agrawal2022context,rubin2022learning,hongjin2022selective}.  
However, related researches about this topic is still limited in our community. As shown in Fig.\ref{fig:motivation} (a), existing works typically randomly select examples~\cite{wang2023images} or simply apply similarity sorting to construct contextual prompts~\cite{zhang2023makes,sun2023exploring}.  
The above insights motivate us to question: 
1) Whether different contexts would significantly impact performance?
2) What is the critical factor of visual prompt selection for ICL-based segmentation models?
In this paper, we first try to answer the above questions. By comprehensive comparisons, we find that the ICL-based models are sensitive to different examples. With multiple experiments, the performance gap between using different contextual examples even exceeds 5 points. Furthermore, we synthetically investigate the effectiveness of the existing prompt selection strategy. Surprisingly, although the similarity measure (\emph{i.e.,} choosing the most similar examples as contexts) provides a relatively effective approach for demonstration search, 
assembling the most dissimilar contexts results in better performance in $\sim$40\% of the test instances. 
To reveal the potential reasons, we empirically demonstrate that the diversity of contexts is a crucial factor in guiding segmentation and enhancing performance (more details in Sec.~\ref{sec:Preliminary}).

Beyond the above discussions, we rethink the role of contextual examples and propose a new \textbf{S}tepwise  \textbf{C}ontext \textbf{S}earch method, which is called \textbf{SCS} for short. As shown in Fig.~\ref{fig:motivation} (b), the SCS focuses on maintaining the diversity of the candidate pool and adaptively searching the well-matched contexts. Meanwhile, our method significantly reduces the annotation cost by searching the typical examples. Specifically, given a large unlabeled database, we first apply a basic cluster algorithm to compact the search space and improve annotation efficiency. Then, we extract the two most dissimilar members in each cluster as candidate examples. By collecting these representative members from all clusters, we construct a small yet rich candidate pool. Furthermore, considering the significant impact of examples on performance, we further introduce a novel adaptive search module based on reinforcement learning. It heuristically utilizes segmentation scores as rewards, guiding the model in determining final demonstrations.   
Extensive experiments across various datasets have shown the effectiveness of our method.
To summarize, our contributions are as follows.
\begin{itemize}
    \item  We take a step forward to systematically analyze the influence of example selection on the ICL-based segmentation frameworks and empirically demonstrate that the diversity of contexts is a key factor in guiding segmentation.
    \item 
    We propose a new SCS method to efficiently select visual prompts by constructing a small yet rich candidate pool. 
    Meanwhile, an adaptive search module is designed to further choose the well-matched demonstrations.
    \item The proposed model significantly improves the segmentation performance of ICL-based frameworks and achieves state-of-the-art on {PASCAL-5$^i$~\cite{shaban2017one}, COCO-20$^i$~\cite{lin2014microsoft} and iSALD-5$^i$~\cite{yao2021scale}.}
\end{itemize}

\section{Related work}
\label{sec:relate}

\subsection{Visual Segmentation}
\label{subsec:visual_seg}
As a fundamental task in computer vision, visual segmentation entails locating and identifying different semantic concepts at the pixel level. 
Numerous works for specific tasks have continued to flourish over the years. For example, Crossover-Net~\cite{yu2021crossover} learns the vertical and horizontal crossover relation for medical image segmentation.
RFNet~\cite{sun2020real} fuses RGB and depth information which is designed for real-time road-driving segmentation.
\cite{hu2020classification} proposes DGEN, a densely connected global entropy network in which noise is suppressed by context information losses.
MSNANet~\cite{rs14194983} accurately extracts water bodies from remote sensing scenes by utilizing multiscale attention. 
Recently, inspired by ICL learning, several general segmentation methods have been proposed~\cite{kirillov2023segment,wang2023images,wang2023SegGPT}. Different from these specialist models, the paradigm can locate visual concepts by feeding a few examples, known as the In-Context Learning (ICL) segmentation frameworks.

\subsection{In-Context Learning}
\label{subsec:in-context}
In-context learning (ICL) is a new learning paradigm that initially emerged from GPT-3~\cite{brown2020language}. It presents a text fill-in-the-blank problem with examples, eliminating the necessity for fine-tuning model parameters on downstream NLP tasks. 
In the vision domain, MAE-VQGAN~\cite{bar2022visual} transforms various visual tasks into grid inpainting problems and proposes the first ICL-based segmentation framework. Different from the previous segmentation methods, this new paradigm provides a promising manner to segment arbitrary objects with a few examples.
Painter~\cite{wang2023images} adopts masked image modeling on continuous pixels, unifying the output form across different visual tasks.
As a variant of Painter, SegGPT~\cite{wang2023SegGPT} integrates visual segmentation tasks into in-context coloring problems and demonstrates powerful capabilities in target segmentation. 
Beyond the previous works, we present comprehensive experiments on example selection for visual ICL models.

\subsection{Contextual Example Selection}
\label{subsec:foundational}
With the development of LLMs, the selection of contextual examples is important for ICL learning~\cite{ye2023compositional,agrawal2022context,rubin2021learning}.
The emerging studies can be categorized into rule-based unsupervised methods and supervised methods. 
KATE~\cite{liu2021makes} uses the KNN mechanism to enhance a retriever for selecting in-context examples. 
EPR~\cite{rubin2021learning} adopts a contrastive learning strategy to train a dense retriever.
UDR~\cite{li2023unified} explores the process of retrieval demonstrations for various tasks in a unified manner.
However, few studies investigate these effects in the vision domain. 
Current works typically select examples randomly~\cite{wang2023images} or simply apply similarity sorting to construct contextual demonstrations~\cite{zhang2023makes,sun2023exploring}. Different from the previous rule-based methods, we introduce an adaptive search method with a small yet rich candidate pool. This strategy effectively reduces the annotation cost and boosts segmentation performance.

\section{Preliminary}
\label{sec:Preliminary}
In this section, we would first present a simple overview of the ICL-based segmentation frameworks. Then, we construct extensive experiments to comprehensively analyze the influence of different examples and the effectiveness of the existing selection strategies. Finally, we experimentally proved that contextual diversity is an important factor for guiding segmentation.

\subsection{Background}
\label{subsec:background}
The design of segmentation models has achieved tremendous development in recent years~\cite{yu2021crossover,zhang2021transfuse,vasudevan2020semantic,li2020reducto}. 
However, these solutions require expensive annotation cost and face challenges in adapting to new scenarios. In contrast, the ICL-based segmentation frameworks~\cite{wang2023SegGPT, bar2022visual} can implement various specific tasks by leveraging several segmented examples. The convenient and versatile vision models provide a promising perspective for our community.

As shown in Fig.~\ref{fig:motivation}, the goal of ICL-based segmentation methods is to predict segmentation regions based on exhibited one or a few prompts. By reorganizing the input format, the models can imitate provided contexts to mask specific object regions. 
Specifically, given a query image $I_q$, these segmentation frameworks first select $n$ example images ${E_d}=\{I_i^{d}\}_{i=1}^{n}$ with corresponding masks ${G_d}=\{G_i^d\}_{i=1}^{n}$ as contextual demonstrations from a huge database. Here, $I_i^d$ and $G_i^d$ denote the $i$-th selected example and corresponding mask. By reformating the input, models can perform multiple segmentation tasks across different scenarios. The above operations can be formulated as:
\begin{equation}\label{goal}
     {\hat{I}_m} = \mathbf{S}({I_q}, {E_d},{G_d}),
\end{equation}
where the ${\textbf{S}}(\cdot)$ and $\hat{I}_m$
are the generalist segmentation model and predicted mask results, respectively. Although the above calculations provide a convenient way to segment objects at arbitrary granularities, the success of the paradigm greatly relies on the given contextual examples. In fact, the basic idea of ICL-based vision models comes from the NLP field, in which it has been proved that contextual prompts play an important role in executing reasoning and boosting performance. However, in the vision community, existing approaches mostly simply choose the demonstrations
(\emph{i.e.,} ${E_d}$ and corresponding ${G_d}$) based on random sampling~\cite{wang2023images} or rely on similarity sorting~\cite{zhang2023makes,sun2023exploring}.  
To explore the impact of the visual prompts on performance, we construct a series of quantitative analyses on the PASCAL dataset~\cite{shaban2017one} with the state-of-the-art ICL-based segmentation model SegGPT~\cite{wang2023SegGPT}. Note that the SegGPT involves two training stages (\emph{i.e.,} pre-training and in-context tuning). In the tuning process, the SegGPT depends on the whole annotation (including base and new classes) to improve the performance. Since the main advantage of the ICL-based framework is to implement few-shot tasks, we only use the \textbf{pre-trained weights} provided by the SegGPT~\cite{wang2023seggpt_simple} to construct experiments.
Following~\cite{bar2022visual,wang2023images}, the mean intersection over union (mIoU) is adopted as the evaluation metric.

\begin{figure}[t]
  \centering
  \begin{subfigure}{0.48\linewidth}
  \hspace{-1mm}
   \includegraphics[width=0.9\linewidth]{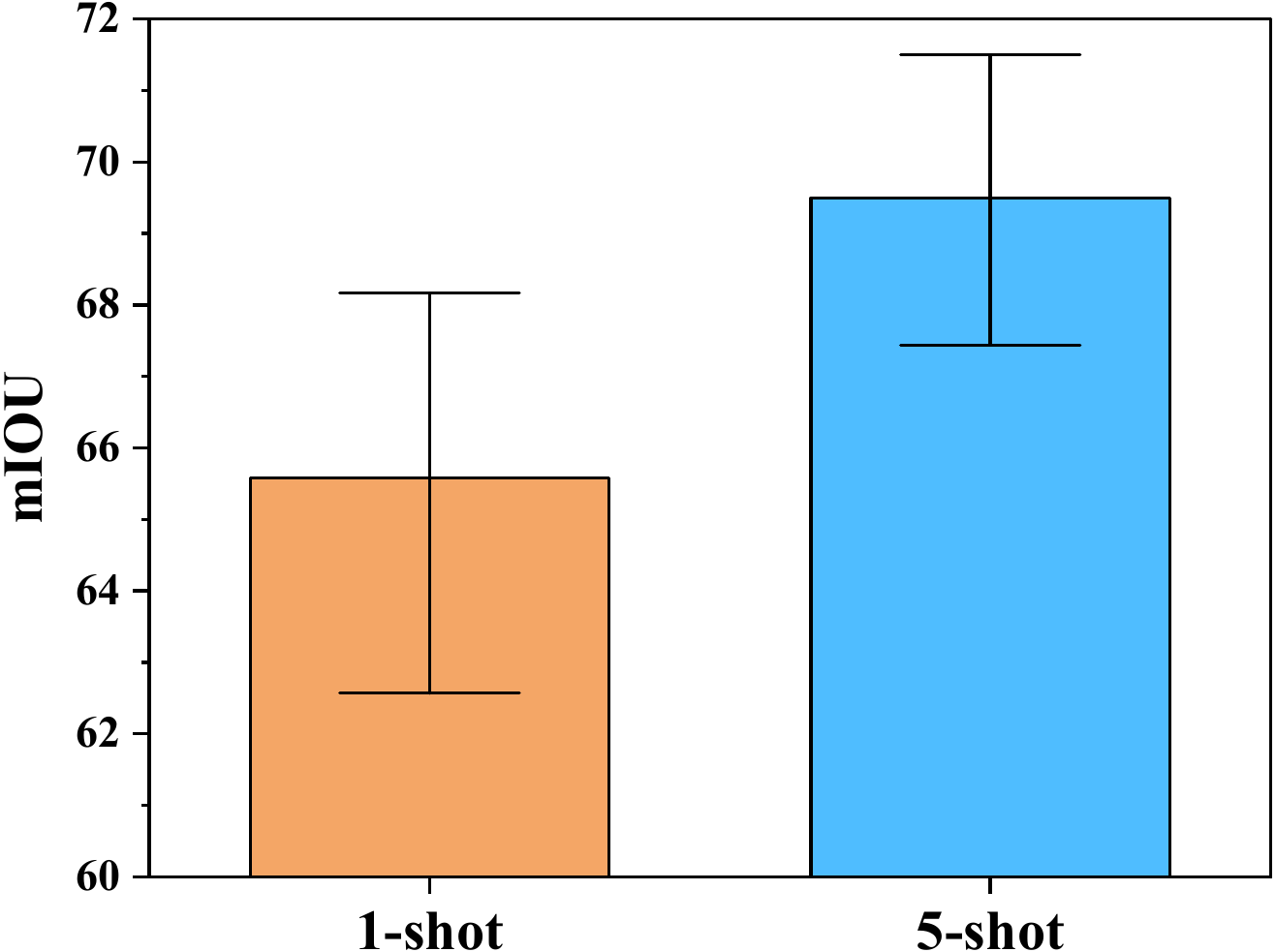}
    \caption{  
    }
    \label{fig:random}
  \end{subfigure}
  \hfill
  \begin{subfigure}{0.5\linewidth}
  \hspace{0mm}
  \includegraphics[width=0.85\linewidth]{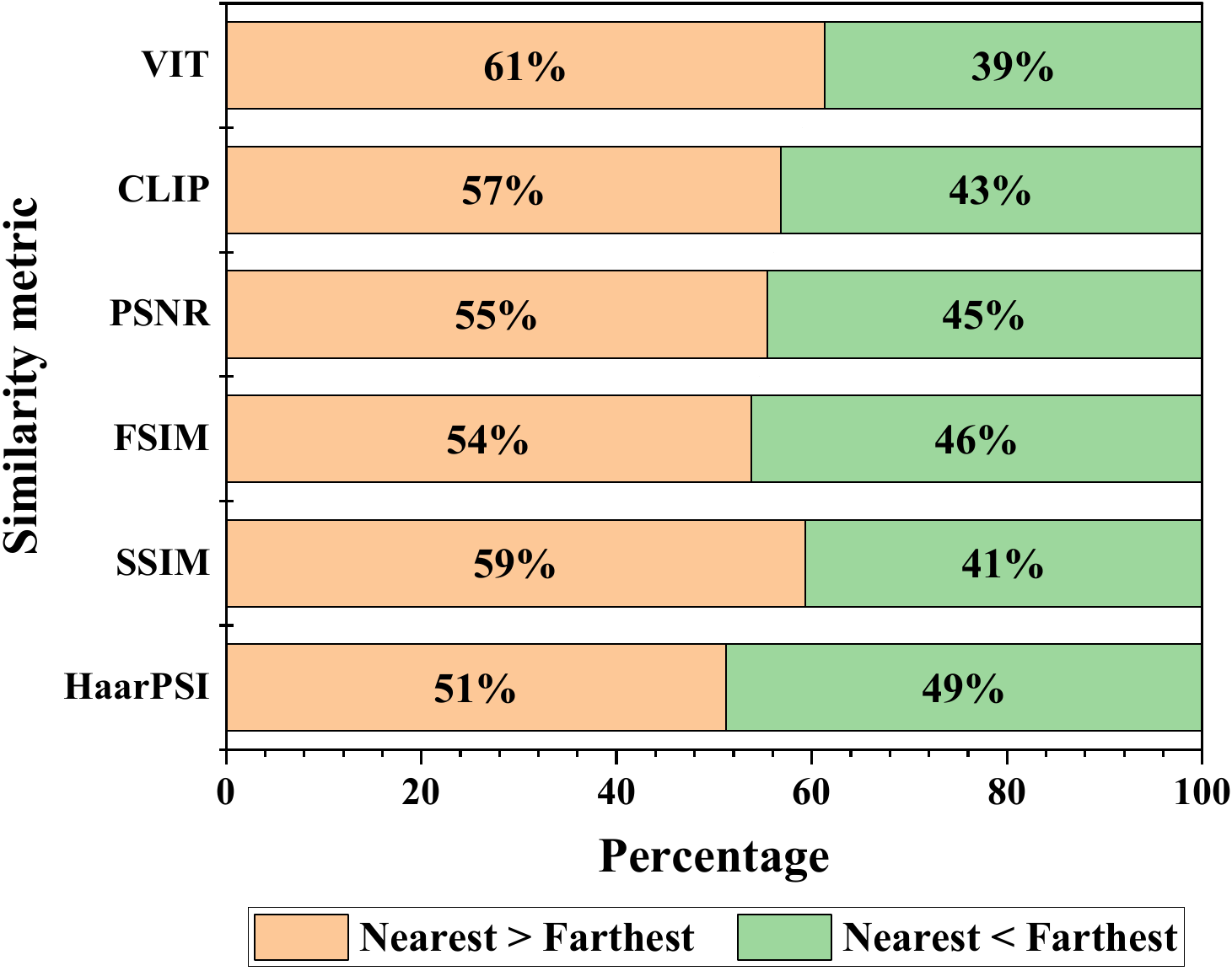}
    \caption{
    }
 \label{fig:similarity}
  \end{subfigure}
  \caption{The influence of different context selection. (a) Randomly sampling contextual examples with 5 runs under 1-shot and 5-shot setting on PASCAL. (b) For each instance, the nearest and farthest examples are retrieved as visual prompts across different similarity-based sorting. Surprisingly, the most dissimilar examples achieved better performance on $\sim$40\% of the test samples.}
  \label{fig:diff_examples}
\end{figure}

\subsection{Influence of Different Examples}
\label{subsec:influence}

As mentioned above, research on the influence of different examples in segmentation remains insufficient. 
In this paper, we aim to fill this gap by conducting experiments to quantitatively analyze the impact of different demonstrations. Specifically, we randomly select examples from the entire training set with 5 different seeds. As shown in Fig.~\ref{fig:diff_examples} (a), we report the best, worst, and average results across the 5 experiments. It reveals that allocating different contexts would dramatically affect the segmentation scores under the 1-shot and 5-shot settings. In particular, the best results can surpass the worst by margins of 5.6 and 4.1 points, respectively. 
Due to the huge performance discrepancy between different contextual prompts, designing an automated method for searching well-matched examples emerges as a critical challenge.


\subsection{Example Selection with Similarity}
\label{subsec:similarity}
To the best of our knowledge, the related researches to our work are~\cite{zhang2023makes,sun2023exploring}, where they simply rely on cosine similarity to select examples with qualitative analysis. In this paper, we try to rethink the strategy through a quantitative discussion across different similarity computation manners. Specifically, we explore two types of image similarity calculation strategies including appearance-level and semantic-level. For the appearance-level comparison, PSNR~\cite{wang2004image}, SSIM~\cite{hore2010image}, FSIM~\cite{zhang2011fsim} and HaarPSI~\cite{reisenhofer2018haar} obtain the similarity between the test instance and the entire training set. 
On the other hand, for semantic-level comparison, we follow~\cite{zhang2023makes,sun2023exploring} to utilize the off-the-shelf image encoders (\emph{i.e.,} CLIP~\cite{radford2021learning} and ViT~\cite{dosovitskiy2020image}) for feature extraction. To verify the effectiveness of these similarity-based approaches, we carefully select the examples based on cosine similarity sorting. 

As shown in Fig.~\ref{fig:diff_examples} (b),
each metric is represented by two bars including $nearest>farthest$ (yellow bar) and $nearest<farthest$ (green bar). 
We take yellow as an example, it
denotes the proportion of instances where using the most similar examples (nearest) as contexts results in a better IoU than using the most dissimilar ones (farthest), and the green bar indicates the opposite. Surprisingly, we observed that assembling the most dissimilar contexts leads to superior performance for numerous test instances. 
There is a substantial portion of the test samples ($\sim$40\%) that achieve better IoU scores by utilizing the farthest examples.
The above results indicate that using similarity measures is not the only way to search for contextual examples. 

\begin{figure}[tb]
  \centering
  \begin{subfigure}{0.48\linewidth}
    \includegraphics[width=0.8\linewidth]{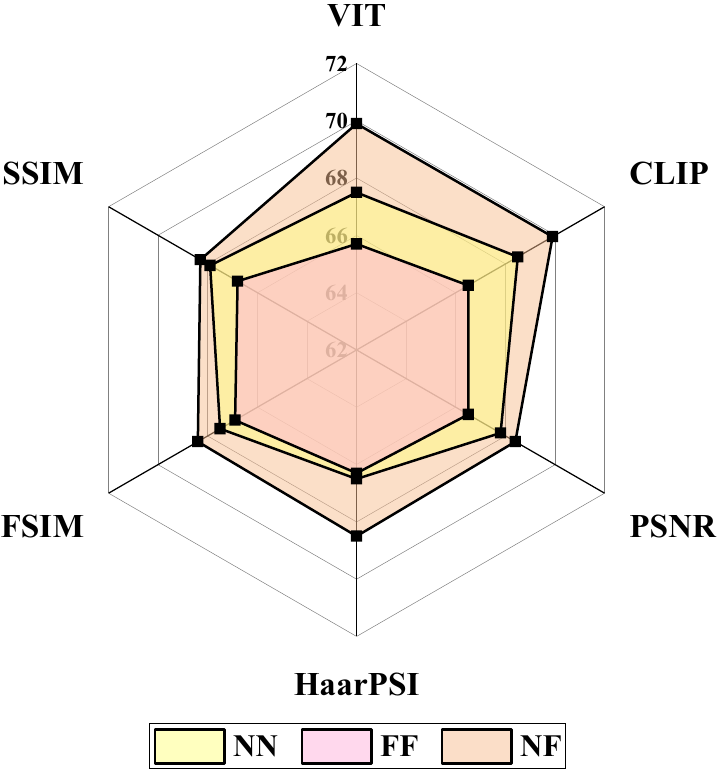}
    \caption{Example sampling with SegGPT~\cite{wang2023SegGPT}.
    }
    \label{example seggpt}
  \end{subfigure}
  \hfill
  \begin{subfigure}{0.5\linewidth}
     \includegraphics[width=0.8\linewidth]{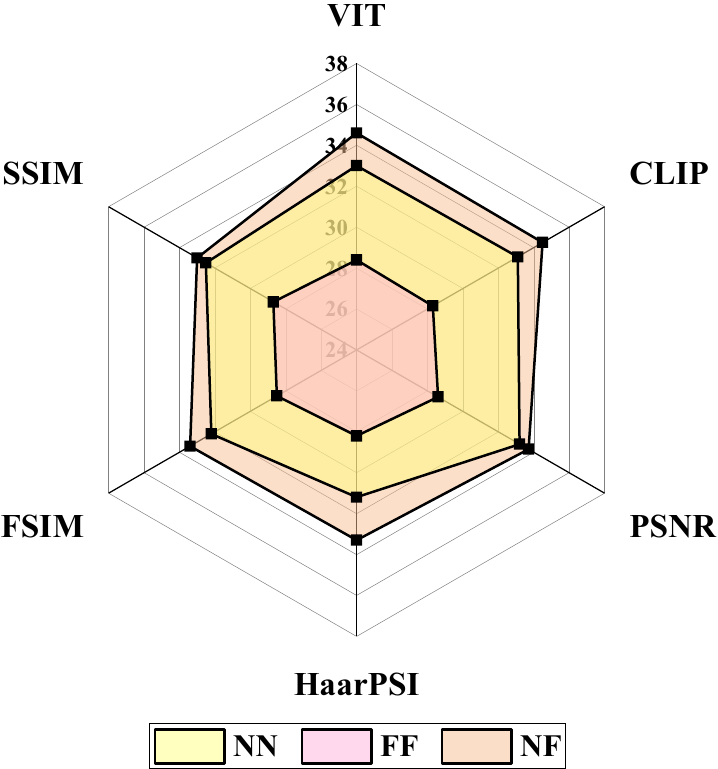}
    \caption{Example sampling with MAE-VQGAN~\cite{bar2022visual}.
    }
 \label{example mae-vqgan}
  \end{subfigure}
  \caption{Diversity \emph{vs} Similarity. Based on similarity sorting, the performance of the two Nearest examples (NN), the two Farthest examples (FF), and the Nearest example with the Farthest example (NF) as visual prompts are shown.}
  \label{fig:diversity}
\end{figure}

\subsection{Example Selection with Diversity}
\label{subsec:diversity}
Based on the above analyses, it is imperative to explore a more effective way for searching visual prompts. 
In Fig.~\ref{fig:diff_examples} (b), we find that applying the most dissimilar examples can also yield satisfactory performance for some instances. The results motivate us to investigate whether combining the nearest and farthest contexts can obtain a performance boost. Consequently, we further construct a series of experiments to verify the hypothesis across different ICL-based models. 

More specifically, based on similarity sorting,
we use two Nearest examples (NN), two Farthest examples (FF), and a combination of the Nearest example and the Farthest example (NF) as contextual demonstrations, respectively. As shown in Fig.~\ref{fig:diversity}, we report related results across different similarity measure manners. It can be observed that the performance using NF examples significantly surpasses the other two selection strategies in all measures. Regarding ViT, the NF achieves better performance that outperforms equipping the two nearest examples by a margin of 2.4 points.


\subsection{Discussion}
\label{subsec:discussion}
In this section, we would try to summarize the above findings with a deeper discussion. Based on the insights presented in Sec.~\ref{subsec:influence}, we find that the ICL-based segmentation framework is sensitive to contextual examples. Given the different visual prompts, the performance gap is even over 5 points on the mIoU. Therefore, building an effective prompt selection method is important for boosting the segmentation results and achieving stable performance. 

To further explore the key factors that influence example selection, we design experiments in Sec.~\ref{subsec:similarity} and Sec.~\ref{subsec:diversity}. The results demonstrate that combining both similar and dissimilar demonstrations can significantly enhance the IoU scores. We argue that the underlying reason is that this manner provides richer semantic information about the segmented objects by expanding the diversity of visual prompts. Based on the above insights, a natural idea is to select these considerable discrepancy examples as contextual prompts for each test instance. However, implementing the straightforward strategy would incur substantial annotation cost and complex computation, since each test sample needs to compare and rerank the candidate list. 
To mitigate the above problems, we introduce a novel stepwise context search (SCS) method that maintains the diversity of the candidate pool and reduces the cost of annotation.
Moreover, considering the necessity of selecting appropriate contextual examples for different test queries, we further design an adaptive search module. It can dynamically select contexts from the candidate pool based on given test samples.

\section{Our Method}
Based on the discussion in Sec.~\ref{sec:Preliminary}, we design a new SCS method to construct a small yet rich candidate pool and adaptively search the contextual demonstrations. 
Compared to previous works, our method has two potential advantages: 1) Fewer Annotations. The SCS focuses on building a small yet diverse candidate pool. It only needs to annotate a portion of samples while achieving better performance. 
2) Adapting Selection. Compared to the previous rule-based mechanism~\cite{zhang2023makes,sun2023exploring}, our method can adaptively search the well-matched examples for different test instances. Next, we would introduce our method in detail.

\subsection{Constructing Candidate Pool}
\label{subsec:CS}
As mentioned in Sec.~\ref{sec:Preliminary}, it is important to provide examples for the ICL-based segmentation paradigm. 
However, previous research mostly depend on huge annotations as contextual candidates, it would inevitably suffers from expensive annotation cost. 
Motivated by the of Sec.~\ref{subsec:discussion}, we aim to build a small yet diverse candidate pool that can be used for subsequent context searching.

Given $N$ unlabeled samples, follow in ~\cite{zhang2023makes}, we utilize CLIP to extract image feature into $d$-dimensional vectors, represented by $F=\{f_i\}_{i=1}^{N}, f_i\in\mathbb{R}^{d}$, where $f_i$ is the $i$-th encoded image feature. 
To construct a candidate pool comprising a set of representative examples,
we employ k-means clustering algorithm~\cite{zhang2022automatic} to generate ${M}$ clusters $C=\{{{c}_m}\}_{m=1}^M$, 
which can be determined by minimizing the sum of squared intra-cluster norms:
\begin{equation}\label{discreteness}
\mathop {\arg\min_C} \sum\limits_{m = 1}^M \sum\limits_{f_i \in {c_j}} {\left\| {{{f_i}} - {{\bf \mu}}_m} \right\|},
\end{equation} 
where ${\mu}_m$ is centroid of cluster ${c_m}$. Initially, this algorithm randomly selects $M$ centroids
. Each instance is assigned by computing the distance to all selected centroids. The process is iteratively executed until the clusters are stable.

To construct the candidate pool, a simple method is to select members closest to the centroid, while previous studies have shown that samples closest to the center do not necessarily represent the entire cluster~\cite{wang2022unsupervised,haussmann2020scalable}. Moreover, related works have provided several alternative sampling methods such as density sorting~\cite{wang2022unsupervised} and entropy sorting~\cite{haussmann2020scalable}.
In this paper, we design a simpler and more efficient sampling strategy without complex computation. Specifically, for
each cluster $c_m$, which includes members $F_m=\{f_k\}_{k=1}^{K}$, where $K$ denotes the number of members in $c_m$. We sort them into a list [$f_k^{(1)}, f_k^{(2)}, ..., f_k^{(K)}$] in the ascending order of their distance to the cluster's center. Based on the findings of Sec.~\ref{subsec:similarity}, aggregating the nearest and farthest examples can cover more diverse demonstrations. Thus, we consider the $f_{mk}^{(1)}$ and $f_{mk}^{(K)}$ as typical samples for each cluster. By collecting these samples from all clusters, we obtain the final candidate pool $D=\{f_{1k}^{(1)},f_{1k}^{(K)},f_{2k}^{(1)},...f_{Mk}^{(1)},f_{Mk}^{(K)}\}$, which contains a total of $2M$ samples. To simplify, we use $f_{d}^{m}$ and $I_d^{m}$ to represent $m$-th image feature and corresponding image, respectively.

\label{sec:ours}

\begin{figure*}[t]
  \centering
   \includegraphics[width=\linewidth]
   {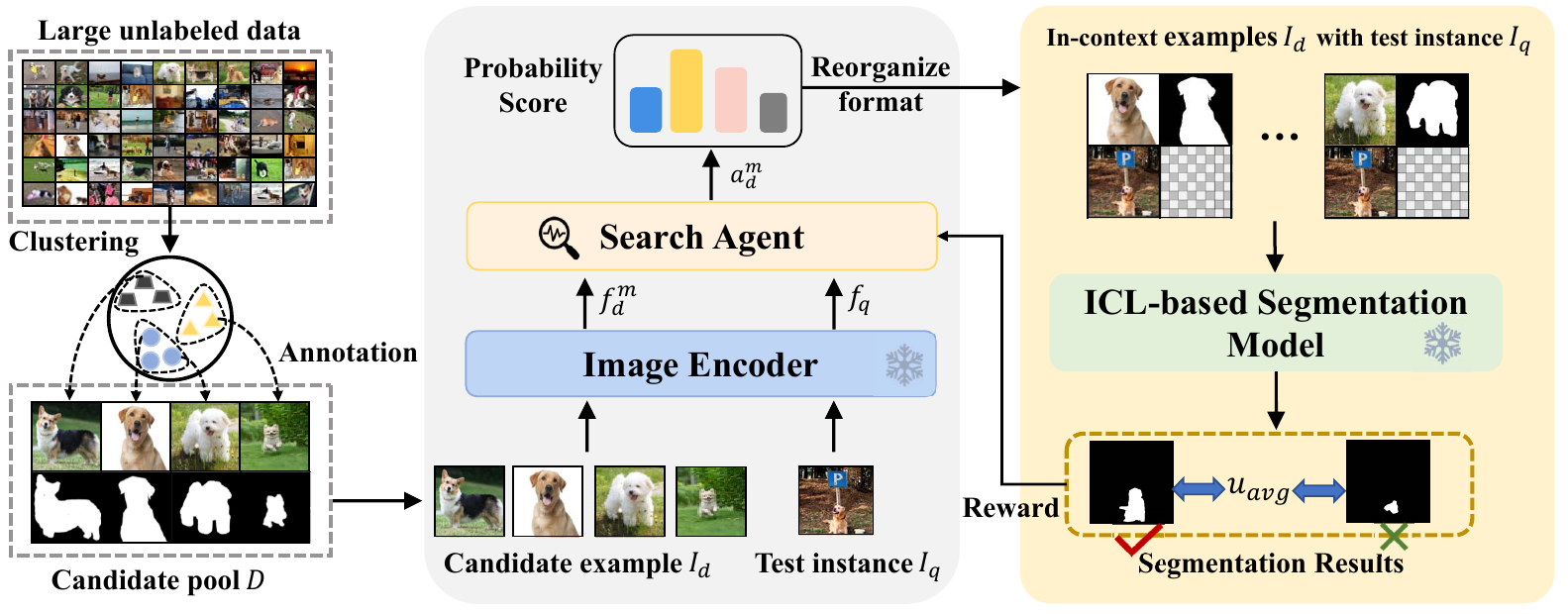}
   \caption{\textbf{Overview of our SCS method.} Instead of similarity sorting on a large annotated dataset, we use clustering to select diverse examples from unlabeled data $D$ and construct the candidate pool. Meanwhile, the search agent is used to further select contextual demonstrations for various test samples based on reinforcement learning. During inference, the test sample and candidate pool examples are fed into the model and adaptively search visual prompts.}
   \label{fig:model}
\end{figure*}

\subsection{Adaptive Search Module}
\label{subsec:reinforce}
In this module, our goal is to select well-matched examples from the constructed candidate pool. 
As shown in Fig.~\ref{fig:model}, we propose a new adaptive search module that can further search and assemble the suitable contexts for given queries. 

Specifically, for a given test instance ${I_q}$, we first apply the same image encoder (\ie CLIP) to extract the corresponding feature vector $f_q\in{\mathbb{R}^d}$. Then, we concatenate the image feature $f_q$ and each example $f_{d}^{m}$ from the candidate pool. These fused features are fed into a search agent that consists of a group of light and simple Multi-Layer Perceptrons (MLP).
The search computation can be denoted as follows:
\begin{align}
\label{projector} 
s_{d}^{m}&= {\mathrm{MLP}}([f_{d}^{m};f_q])), \\
\label{softmax}
a_{d}^{1},a_{d}^{2},...,a_{d}^{2M}&= \mathrm{softmax}(s_{d}^{1},s_{d}^{2},...,s_{d}^{2M}),
\end{align}
where $a_{d}^{m}$ represents the probability of $m$-th example in the candidate pool. 
To guide the search agent in learning select examples, we apply the IoU scores as rewards for selecting well-matched demonstrations from the candidate pool. Specifically, following the format of the ICL-based segmentation model, we reorganize query $I_q$, the original image of context examples $I_d$ and corresponding mask $G_d$ into a joint image, as shown in Fig.~\ref{fig:model}. These images are then fed into the ICL-based segmentation model to obtain the IoU scores $U=\{u_d\}_{d=1}^{2M}$. Inspired by~\cite{anderson2018bottom,suo2023s3c}, we use the average IoU score $u_{avg}$ as the base score. By applying reinforcement learning, the gradient is calculated by:
\begin{align}
   \nabla _\theta L(\theta) = -\frac{1}{2M} \sum\limits_{m=1}^{2M}(u_d^{m}-u_{avg})\nabla _\theta log (a_d^{m}),
\end{align}
where $a_d^{m}$ and $\theta$ are the probability of $m$-th example and the parameters of the search agent. Based on the above computation, this gradient would tend to increase
the probability of the $m$-th context when its IoU scores $u_d^{m}$ are higher than the average IoU score $u_{avg}$. 

During inference, we directly input the test samples and candidate pool examples into the search agent. The contextual examples are then determined by sorting them based on probability scores $a_d$  as detailed in Eq.~\ref{softmax}.

\section{Experiment}
\label{sec:experiment}
We use different prompt selection strategies and various generalist segmentation frameworks to demonstrate the superiority of our method. Furthermore, we compare our approach with traditional methods and conduct extensive qualitative and quantitative studies on multiple datasets.

\begin{table}[t!]
\small
\begin{center}
\caption{The results on COCO-20$^i$ and PASCAL-5$^i$ based on example-based semantic segmentation. To ensure a fair comparison and follow the few-shot setting, 
$\dagger$ indicates that only using the \textbf{pre-trained weights} provided by the SegGPT~\cite{wang2023seggpt_simple}, and we report related results based on randomly sampling (\emph{i.e.,} w/R) and similarity sorting~\cite{zhang2023makes}.}
\footnotesize{
\setlength{\tabcolsep}{1pt}
\begin{tabular}{ccccccccc} 
 \toprule[1pt]
 &\multicolumn{4}{c}{COCO-20$^i$~\cite{lin2014microsoft}} & \multicolumn{4}{c}{PASCAL-5$^i$~\cite{shaban2017one}} \\
 \cmidrule(lr){2-5}\cmidrule(lr){6-9}
 $\mathbf{Method}$ & \multicolumn{2}{c}{1-shot}  & \multicolumn{2}{c}{5-shot} & \multicolumn{2}{c}{1-shot}  & \multicolumn{2}{c}{5-shot}  \\
 \cmidrule(lr){2-3} \cmidrule(lr){4-5}\cmidrule(lr){6-7}\cmidrule(lr){8-9}
     & mIoU & FBIoU & mIoU & FBIoU & mIoU & FBIoU & mIoU & FBIoU   \\
        \midrule
        Painter~\cite{wang2023images}  & 32.8 & 50.2 & 32.6 & 50.1 & 64.5 & 74.2 & 64.6 & 74.4 \\
        MAE-VQGAN~\cite{bar2022visual}  & 12.1 & 45.7 & - & - & 27.5 & 55.6 & - & - \\
        MAE-VQGAN w/~\cite{zhang2023makes}  & 18.0 & 46.5 & - & - & 33.6 & 57.8 & - & - \\
        SegGPT$\dagger$ w/R~\cite{wang2023SegGPT} & 48.6 & 71.3 & 52.7 & 73.1 & 72.1 & 83.1 & 76.2 & 80.3 \\
        SegGPT$\dagger$ w/~\cite{zhang2023makes}  & 53.3 & 72.2 & 56.1 & 75.5 & 74.4 & 84.4 & 76.9 & 85.9 \\
        \hline
        MAE-VQGAN w/Ours & 19.1 & 47.0 & - & - & 35.0 & 61.2 & - & - \\
        SegGPT$\dagger$ w/Ours & \textbf{57.2} & \textbf{75.1} & \textbf{59.5} & \textbf{76.8} & \textbf{76.6} & \textbf{85.4} & \textbf{78.2} & \textbf{86.5} \\
        \bottomrule[1pt]
\end{tabular}

}
\label{tab:main_results}
\end{center}

\end{table}

\subsection{Experimental Setting}
\label{subsec:setting}
\textbf{Dataset.} 
We follow the previous works~\cite{bar2022visual,zhang2023makes,wang2023SegGPT} and evaluate our approach on PASCAL-5$^i$~\cite{shaban2017one}, COCO-20$^i$~\cite{lin2014microsoft} and iSALD-5$^i$~\cite{yao2021scale}.
PASCAL-5$^i$ includes 20 semantic categories and is created from PASCAL VOC 2012~\cite{everingham2010pascal} with external annotations from SDS~\cite{hariharan2011semantic}. 
COCO-20$^i$ is built upon MSCOCO~\cite{lin2014microsoft}, which is a more challenging dataset and contains 80 categories. For all categories, both COCO-20$^i$ and PASCAL-5$^i$ are evenly divided into 4 folds for cross-validation, where each fold is selected as the test set and the remaining three folds are used for training. The
iSALD-5$^i$ contains 15 geospatial categories which derived from iSALD~\cite{waqas2019isaid} in the remote sensing image semantic segmentation task.
Similarly, the iSALD-5$i$ is also partitioned into 3 folds for cross-validation.

\noindent\textbf{Evaluation Metric.} Following~\cite{hong2022cost}, mean intersection over union (mIoU) and foreground-background IoU (FBIoU) are adopted as evaluation metrics. The former measures the average overall IoU values for all object classes, while the latter reports the average IoU scores for the foreground and background. 

\noindent\textbf{Implementation details.} 
For both datasets, the Adam~\cite{kingma2014adam} is used as our optimizer and the maximum number of training epochs is 10.
We use the CLIP-ViT~\cite{radford2021learning} image encoder to extract image features.
During training, all images are resized to 224×224 and fed into the image encoder. The dimension of extracted feature vectors is 1024.
We train on one 4080 GPU with an initial learning rate of 1e-3, which is dropped by half every 5 epochs. The number of clusters $M$ and batch size is set to 10 and 8.

\subsection{Quantitative Evaluation}
\label{subsec:results}
        
        
        
        
        
       

In Table~\ref{tab:main_results}, we show a performance comparison of two general ICL-based segmentation models
with and without proposed SCS. Following~\cite{zhang2023makes,sun2023exploring}, all results are reported by repeating the random experiment five times.
It can be observed that our method can significantly boost performance in all settings including 1-shot and 5-shot settings.
In particular, for the 1-shot setting, the SCS leads to MAE-VQGAN with the mIoU gain of 7.0 and 6.5 on the PASCAL-5$^i$ and COCO-20$^i$ dataset, respectively.
Besides, compared to previous rule-base models~\cite{zhang2023makes} in 1-shot and 5-shot semantic segmentation tasks, our method surpasses it by a considerable margin. Moreover, 
following the few-shot setting and ensuring a fair comparison, we report related results based on the pre-trained weights for SegGPT~\cite{wang2023seggpt_simple}.
It can be observed that our method significantly outperforms these rule-based visual prompt selection methods for SegGPT by 9.7 and 6.8 on the 1-shot and 5-shot settings for COCO-20$^i$. 
More importantly, the above results demonstrate that the proposed SCS can be considered as a ``plug-and-play'' module to enhance the segmentation ability of ICL-based frameworks.

\subsection{Ablation Study}
\label{subsec:ablation}
To clarify the effectiveness of our method, we conduct further ablation studies on the COCO-20$^i$ dataset with the 5-shot setting. 
The SegGPT is used as the basic model due to its outstanding segmentation performance. 

\begin{table}[t!]
\caption{\textbf{Ablation study.} \textbf{Similarity:} Cosine similarity ranking; \textbf{Cluster:} Cluster compacting; \textbf{Sorting:} The nearest-farthest sampling; \textbf{ASM:} Adaptive search module.} 
\centering
\begin{center}
\footnotesize{
\setlength{\tabcolsep}{5pt}
\begin{tabular}{c|cccc|cc}
 \hline
 ~ &Similarity & Cluster & Sorting & ASM & mIoU & FBIoU \\
        \hline
         1 &  &  &  &  &52.7 & 73.1 \\
         \hline
        2 &\checkmark &  &  &  & 56.1& 74.5 \\
        3 & \checkmark& \checkmark &  &  & 55.7& 74.3 \\
        4 &\checkmark& \checkmark & \checkmark &  & 57.6& 75.7  \\
        5 & & \checkmark & \checkmark & \checkmark & \textbf {59.5}& \textbf{76.8} \\
        \hline
\end{tabular}
}
\end{center}
\label{tab:Ablation_studies}
\end{table}

\noindent\textbf{Effects of Cluster.}
In Table~\ref{tab:Ablation_studies}, we first build a baseline by randomly selecting examples from the entire training set. The average result is reported across 5 different seeds. In the second row of Table~\ref{tab:Ablation_studies}, we follow~\cite{zhang2023makes} and utilize cosine similarity ranking to select the top-5 examples as contexts. In the third row, we investigate the impact of the cluster on in-context example selection, using the closest examples to each cluster centroid to construct the candidate pool. Compared to the baseline, the results show that compacting the search space not only reduces annotation cost but also improves model performance. 
\begin{table}[!t]
     \caption{
     The effects of different settings for our method.
     }
    \centering
    \small
      \begin{tabular}{cc|cccc}
        \hline
         &\multirow{2}{*}{Model}& \multicolumn{2}{c}{1-shot} & 
        \multicolumn{2}{c}{5-shot} \\
        
        
        && {mIoU}  & {FBIoU} & {mIoU}  & {FBIoU}  \\
        
         \hline
        1&Entropy & 53.1 & 72.3 & 56.1 & 75.0 \\
        2&Foreground & 53.5 & 72.5 & 56.7 & 79.3 \\
        \hline
        3&ViT & 56.9 & 75.0 & 58.7 & 76.0 \\
        4&CLIP-resnet & 56.3 & 74.7 & 58.5 & 76.4 \\
        \hline
        5&Density-sort& \textbf{57.6} & 75.1 & 59.1 & 76.3 \\
        6&Entropy-sort& 56.9 & 74.5 & 58.4 & 75.9 \\
        \hline
        7&Ours& 57.2 & \textbf{75.1} & \textbf{59.5} & \textbf{76.8} \\
       \hline
      \end{tabular}
    \label{tab:distance}
\end{table}

\noindent\textbf{Effects of Sort.}
As shown in row 4 of Table~\ref{tab:Ablation_studies}, we experimentally demonstrate that integrating our sorting mechanism results in 1.9 mIoU improvement compared to directly using the closest examples to each cluster centroid. 
The inherent reason is that centroid examples describe a limited feature space~\cite{wang2022unsupervised,shen2020large}, making it challenging to support a variety of different contexts for test instances.  
In contrast, our sorting mechanism introduces the nearest and farthest examples 
, which constructs more diverse demonstrations for ICL learning.

\noindent\textbf{Effects of Adaptive Search Module.}
Finally, we further explore the effectiveness of our adaptive search module in the last row of Table~\ref{tab:Ablation_studies}. Compared to using a simple similarity sorting method, our approach shows an improvement of 3.4 points. Meanwhile, benefiting from the diverse candidate pool, our SCS achieves state-of-the-art performance and outperforms the baseline by a considerable margin.

\begin{table}[t]
\setlength{\tabcolsep}{3pt}
    \parbox{.48\linewidth}{
    \centering
\caption{The results on PASCAL-5$^i$~\cite{shaban2017one} based on instance segmentation.}
\label{tab:instance_seg}
\begin{tabular}{c|cc}
\hline
  {Model}& {1-shot} & {5-shot} \\
  \hline
  SegGPT$\dagger$ w/R~\cite{wang2023SegGPT} & 46.5 & 46.9  \\
  SegGPT$\dagger$ w/~\cite{zhang2023makes} & 47.1 & 48.5   \\
  SegGPT w/Ours & \textbf{50.5} & \textbf{51.4}   \\
\hline
\end{tabular}}
\parbox{.48\linewidth}{
\centering
\caption{The results on iSALD-5$^i$~\cite{waqas2019isaid} based on semantic segmentation.}
\label{tab:aerial_seg}
\begin{tabular}{c|cc}
\hline
  {Model}& {1-shot} & {5-shot} \\
  \hline
  SegGPT$\dagger$ w/R~\cite{wang2023SegGPT} & 29.8 & 30.6  \\
  SegGPT$\dagger$ w/~\cite{zhang2023makes} & 34.7 & 35.4   \\
  SegGPT w/Ours & \textbf{37.9} & \textbf{39.5}   \\
 \hline
\end{tabular}
}
\end{table}

\subsection{Different Model Settings}
In this section, we explore several alternative model settings to further discuss the proposed method. All experiments are conducted on the COCO-20$^i$ with the SegGPT model. In the first two rows, we first compare the performance using different metrics. 
Specifically, we use the rule-based methods to select the contextual examples from the complete training set. which including Entropy sorting~\cite{haussmann2020scalable} and Foreground sorting~\cite{yu2023h2rbox}. 
In particular, ``Entropy'' is executed by calculating the maximum entropy values. The ``Foreground'' refers to a two-stage approach~\cite{yu2023h2rbox}. It uses an object detector~\cite{wang2023internimage} to extract 
the foreground boxes and the foreground similarity is used as the metric. From the 1-shot and 5-shot results, it can be observed that our SCS can dramatically outperform these methods by a large margin.
In rows 3-4, we construct the experiments with different feature encoders. We replace the CLIP-ViT with ViT~\cite{dosovitskiy2020image} and CLIP-resnet~\cite{radford2021learning} to obtain image feature vectors, These extracted features would be utilized to construct the candidate pool and perform the adaptive searching. The results show that our SCS is robust for different image encoders.
Finally, in rows 5-6 of Table 3, we replace our selection mechanism with several alternative manners. However, these approaches mostly involve complex computation. For example, the Density-sort~\cite{wang2022unsupervised_simple} requires additional computations on the order of $\boldsymbol{O(MN)}$ in each iteration to update density lists, where $M$ is the number of clusters and $N$ is the size of the unlabeled data. Instead, our method uses the basic clustering algorithm and obtains a comparable performance.

\subsection{Further Discussion}

\noindent\textbf{More Segmentation tasks.}
In this section, we evaluate the effect of our method on more segmentation tasks and datasets. Because MAE-VQGAN cannot distinguish between individual instances of objects within the same class, we construct more experiments with the SegGPT model. Specifically, we report instance segmentation results on PASCAL-5i in Table~\ref{tab:instance_seg} and aerial image semantic segmentation results on iSALD-5i in Table~\ref{tab:aerial_seg}.
The above results indicate that the SCS can enhance the segmentation ability for various real-world scenarios.

\begin{figure*}[t]
  \centering
   \includegraphics[width=0.83\linewidth]{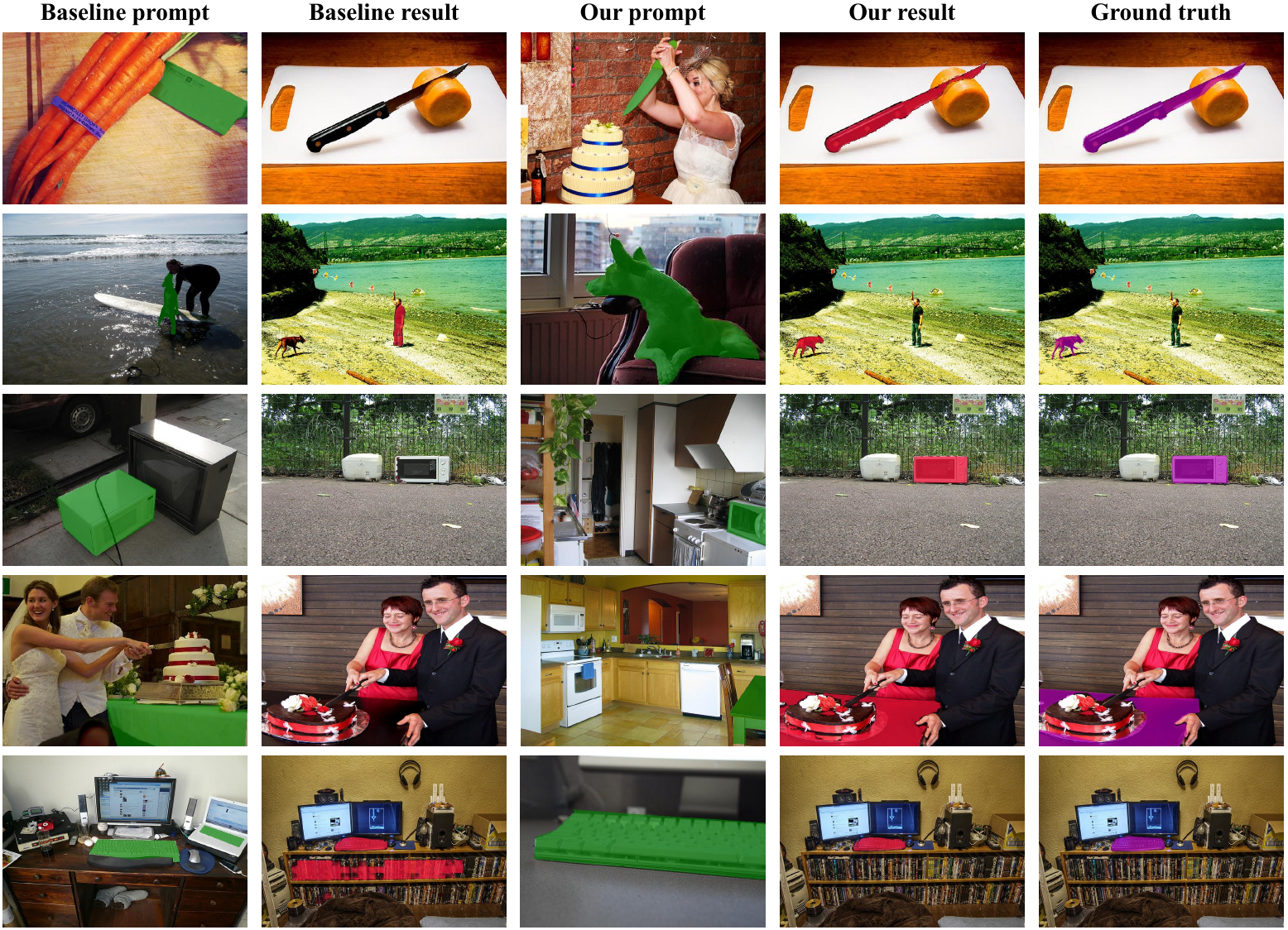}
   \caption{
   Qualitative results under 1-shot setting on COCO-$20^i$. The similarity-based selection method is viewed as our baseline. The green regions, red regions and purple regions are example masks, predicted masks and ground-truth masks.
   }
   \label{fig:qual_results}
\end{figure*}

\noindent\textbf{Qualitative Results.}
As shown in Fig.~\ref{fig:qual_results}, to further analyze and verify the proposed method, we visualized the qualitative segmentation results on COCO-20$^i$ with the SegGPT. 
The first and third columns represent the visual prompts (highlighted in green) based on similarity sorting and our SCS method, respectively. 
The second and fourth columns show the corresponding segmentation results (highlighted in red).
The last column represents the ground truth mask (highlighted in purple).
As illustrated in the first row, the baseline model selects the most similar context to the query (\textit{i.e.,} chopping board, knife and carrot), while the segment result is not satisfactory, as the model can not recognize the knife in the image. Instead, Our method selects an image from the candidate pool that accurately segments out the mask. The above results demonstrate that similarity sorting is not the only golden standard and improving the diversity of context prompts can boost the segmentation results.

\section{Conlusion}
\label{sec:conlusion}
In this paper, we focus on the example selection method for the ICL-based segmentation frameworks. By building comprehensive comparisons, 
we empirically demonstrate that diversity of contexts is a key factor in guiding segmentation and boosting performance. 
Beyond the above insights, we propose a new stepwise context search method that improves the diversity of the candidate pool and significantly alleviates the annotation cost.  
 We hope that this work can promote further research to better understand in-context learning in computer vision.

\clearpage  

%
%
\noindent\textbf{Acknowledgement.} This work was supported by National Natural Science Foundation of
China (No.62102323), the Innovation Capability Support Program of Shaanxi(Program No. 2023KJXX-142) and National Natural Science Foundation of
China (U23B2013).

\bibliographystyle{splncs04}
\bibliography{main}

\begin{thebibliography}{10}
\providecommand{\url}[1]{\texttt{#1}}
\providecommand{\urlprefix}{URL }
\providecommand{\doi}[1]{https://doi.org/#1}

\bibitem{agrawal2022context}
Agrawal, S., Zhou, C., Lewis, M., Zettlemoyer, L., Ghazvininejad, M.: In-context examples selection for machine translation. arXiv preprint arXiv:2212.02437  (2022)

\bibitem{wang2022unsupervised_simple}
et~al., W.: "unsupervised selective labeling for more effective semi-supervised learning". In: ECCV 2022. pp. 427--445. Springer (2022)

\bibitem{wang2023seggpt_simple}
et~al, W.: "seggpt: Segmenting everything in context". In ICCV 2023  (2023)

\bibitem{anderson2018bottom}
Anderson, P., He, X., Buehler, C., Teney, D., Johnson, M., Gould, S., Zhang, L.: Bottom-up and top-down attention for image captioning and visual question answering. In: Proceedings of the IEEE conference on computer vision and pattern recognition. pp. 6077--6086 (2018)

\bibitem{badrinarayanan2017segnet}
Badrinarayanan, V., Kendall, A., Cipolla, R.: Segnet: A deep convolutional encoder-decoder architecture for image segmentation. IEEE transactions on pattern analysis and machine intelligence  \textbf{39}(12),  2481--2495 (2017)

\bibitem{bar2022visual}
Bar, A., Gandelsman, Y., Darrell, T., Globerson, A., Efros, A.: Visual prompting via image inpainting. Advances in Neural Information Processing Systems  \textbf{35},  25005--25017 (2022)

\bibitem{brown2020language}
Brown, T., Mann, B., Ryder, N., Subbiah, M., Kaplan, J.D., Dhariwal, P., Neelakantan, A., Shyam, P., Sastry, G., Askell, A., et~al.: Language models are few-shot learners. Advances in neural information processing systems  \textbf{33},  1877--1901 (2020)

\bibitem{dosovitskiy2020image}
Dosovitskiy, A., Beyer, L., Kolesnikov, A., Weissenborn, D., Zhai, X., Unterthiner, T., Dehghani, M., Minderer, M., Heigold, G., Gelly, S., et~al.: An image is worth 16x16 words: Transformers for image recognition at scale. arXiv preprint arXiv:2010.11929  (2020)

\bibitem{everingham2010pascal}
Everingham, M., Van~Gool, L., Williams, C.K., Winn, J., Zisserman, A.: The pascal visual object classes (voc) challenge. International journal of computer vision  \textbf{88},  303--338 (2010)

\bibitem{hariharan2011semantic}
Hariharan, B., Arbel{\'a}ez, P., Bourdev, L., Maji, S., Malik, J.: Semantic contours from inverse detectors. In: 2011 international conference on computer vision. pp. 991--998. IEEE (2011)

\bibitem{haussmann2020scalable}
Haussmann, E., Fenzi, M., Chitta, K., Ivanecky, J., Xu, H., Roy, D., Mittel, A., Koumchatzky, N., Farabet, C., Alvarez, J.M.: Scalable active learning for object detection. In: 2020 IEEE intelligent vehicles symposium (iv). pp. 1430--1435. IEEE (2020)

\bibitem{he2017mask}
He, K., Gkioxari, G., Doll{\'a}r, P., Girshick, R.: Mask r-cnn. In: Proceedings of the IEEE international conference on computer vision. pp. 2961--2969 (2017)

\bibitem{hong2022cost}
Hong, S., Cho, S., Nam, J., Lin, S., Kim, S.: Cost aggregation with 4d convolutional swin transformer for few-shot segmentation. In: European Conference on Computer Vision. pp. 108--126. Springer (2022)

\bibitem{hongjin2022selective}
Hongjin, S., Kasai, J., Wu, C.H., Shi, W., Wang, T., Xin, J., Zhang, R., Ostendorf, M., Zettlemoyer, L., Smith, N.A., et~al.: Selective annotation makes language models better few-shot learners. In: The Eleventh International Conference on Learning Representations (2022)

\bibitem{hore2010image}
Hore, A., Ziou, D.: Image quality metrics: Psnr vs. ssim. In: 2010 20th international conference on pattern recognition. pp. 2366--2369. IEEE (2010)

\bibitem{hu2020classification}
Hu, H., Li, Z., Li, L., Yang, H., Zhu, H.: Classification of very high-resolution remote sensing imagery using a fully convolutional network with global and local context information enhancements. IEEE Access  \textbf{8},  14606--14619 (2020)

\bibitem{kingma2014adam}
Kingma, D.P., Ba, J.: Adam: A method for stochastic optimization. arXiv preprint arXiv:1412.6980  (2014)

\bibitem{kirillov2023segment}
Kirillov, A., Mintun, E., Ravi, N., Mao, H., Rolland, C., Gustafson, L., Xiao, T., Whitehead, S., Berg, A.C., Lo, W.Y., et~al.: Segment anything. arXiv preprint arXiv:2304.02643  (2023)

\bibitem{li2023unified}
Li, X., Lv, K., Yan, H., Lin, T., Zhu, W., Ni, Y., Xie, G., Wang, X., Qiu, X.: Unified demonstration retriever for in-context learning. arXiv preprint arXiv:2305.04320  (2023)

\bibitem{li2020reducto}
Li, Y., Padmanabhan, A., Zhao, P., Wang, Y., Xu, G.H., Netravali, R.: Reducto: On-camera filtering for resource-efficient real-time video analytics. In: Proceedings of the Annual conference of the ACM Special Interest Group on Data Communication on the applications, technologies, architectures, and protocols for computer communication. pp. 359--376 (2020)

\bibitem{lin2014microsoft}
Lin, T.Y., Maire, M., Belongie, S., Hays, J., Perona, P., Ramanan, D., Doll{\'a}r, P., Zitnick, C.L.: Microsoft coco: Common objects in context. In: Computer Vision--ECCV 2014: 13th European Conference, Zurich, Switzerland, September 6-12, 2014, Proceedings, Part V 13. pp. 740--755. Springer (2014)

\bibitem{liu2021makes}
Liu, J., Shen, D., Zhang, Y., Dolan, B., Carin, L., Chen, W.: What makes good in-context examples for gpt-$3 $? arXiv preprint arXiv:2101.06804  (2021)

\bibitem{long2015fully}
Long, J., Shelhamer, E., Darrell, T.: Fully convolutional networks for semantic segmentation. In: Proceedings of the IEEE conference on computer vision and pattern recognition. pp. 3431--3440 (2015)

\bibitem{luo2020multi}
Luo, G., Zhou, Y., Sun, X., Cao, L., Wu, C., Deng, C., Ji, R.: Multi-task collaborative network for joint referring expression comprehension and segmentation. In: Proceedings of the IEEE/CVF Conference on computer vision and pattern recognition. pp. 10034--10043 (2020)

\bibitem{rs14194983}
Lyu, X., Fang, Y., Tong, B., Li, X., Zeng, T.: Multiscale normalization attention network for water body extraction from remote sensing imagery. Remote Sensing  \textbf{14}(19) (2022). \doi{10.3390/rs14194983}, \url{https://www.mdpi.com/2072-4292/14/19/4983}

\bibitem{min2022rethinking}
Min, S., Lyu, X., Holtzman, A., Artetxe, M., Lewis, M., Hajishirzi, H., Zettlemoyer, L.: Rethinking the role of demonstrations: What makes in-context learning work? arXiv preprint arXiv:2202.12837  (2022)

\bibitem{radford2021learning}
Radford, A., Kim, J.W., Hallacy, C., Ramesh, A., Goh, G., Agarwal, S., Sastry, G., Askell, A., Mishkin, P., Clark, J., et~al.: Learning transferable visual models from natural language supervision. In: International conference on machine learning. pp. 8748--8763. PMLR (2021)

\bibitem{reisenhofer2018haar}
Reisenhofer, R., Bosse, S., Kutyniok, G., Wiegand, T.: A haar wavelet-based perceptual similarity index for image quality assessment. Signal Processing: Image Communication  \textbf{61},  33--43 (2018)

\bibitem{ronneberger2015u}
Ronneberger, O., Fischer, P., Brox, T.: U-net: Convolutional networks for biomedical image segmentation. In: Medical Image Computing and Computer-Assisted Intervention--MICCAI 2015: 18th International Conference, Munich, Germany, October 5-9, 2015, Proceedings, Part III 18. pp. 234--241. Springer (2015)

\bibitem{rubin2021learning}
Rubin, O., Herzig, J., Berant, J.: Learning to retrieve prompts for in-context learning. arXiv preprint arXiv:2112.08633  (2021)

\bibitem{rubin2022learning}
Rubin, O., Herzig, J., Berant, J.: Learning to retrieve prompts for in-context learning. In: Proceedings of the 2022 Conference of the North American Chapter of the Association for Computational Linguistics: Human Language Technologies. pp. 2655--2671 (2022)

\bibitem{shaban2017one}
Shaban, A., Bansal, S., Liu, Z., Essa, I., Boots, B.: One-shot learning for semantic segmentation. arXiv preprint arXiv:1709.03410  (2017)

\bibitem{shen2020large}
Shen, F., Zhao, L., Du, W., Zhong, W., Qian, F.: Large-scale industrial energy systems optimization under uncertainty: A data-driven robust optimization approach. Applied Energy  \textbf{259},  114199 (2020)

\bibitem{shi2018key}
Shi, H., Li, H., Meng, F., Wu, Q.: Key-word-aware network for referring expression image segmentation. In: Proceedings of the European Conference on Computer Vision (ECCV). pp. 38--54 (2018)

\bibitem{sun2020real}
Sun, L., Yang, K., Hu, X., Hu, W., Wang, K.: Real-time fusion network for rgb-d semantic segmentation incorporating unexpected obstacle detection for road-driving images. IEEE robotics and automation letters  \textbf{5}(4),  5558--5565 (2020)

\bibitem{sun2024adaptive}
Sun, M., Suo, W., Wang, P., Niu, K., Liu, L., Lin, G., Zhang, Y., Wu, Q.: An adaptive correlation filtering method for text-based person search. International Journal of Computer Vision pp. 1--16 (2024)

\bibitem{sun2023exploring}
Sun, Y., Chen, Q., Wang, J., Wang, J., Li, Z.: Exploring effective factors for improving visual in-context learning. arXiv preprint arXiv:2304.04748  (2023)

\bibitem{suo2023s3c}
Suo, W., Sun, M., Liu, W., Gao, Y., Wang, P., Zhang, Y., Wu, Q.: S3c: Semi-supervised vqa natural language explanation via self-critical learning. In: Proceedings of the IEEE/CVF Conference on Computer Vision and Pattern Recognition. pp. 2646--2656 (2023)

\bibitem{suo2022simple}
Suo, W., Sun, M., Niu, K., Gao, Y., Wang, P., Zhang, Y., Wu, Q.: A simple and robust correlation filtering method for text-based person search. In: European conference on computer vision. pp. 726--742. Springer (2022)

\bibitem{suo2022rethinking}
Suo, W., Sun, M., Wang, P., Zhang, Y., Wu, Q.: Rethinking and improving feature pyramids for one-stage referring expression comprehension. IEEE Transactions on Image Processing  \textbf{32},  854--864 (2022)

\bibitem{vasudevan2020semantic}
Vasudevan, A.B., Dai, D., Van~Gool, L.: Semantic object prediction and spatial sound super-resolution with binaural sounds. In: European conference on computer vision. pp. 638--655. Springer (2020)

\bibitem{wang2023internimage}
Wang, W., Dai, J., Chen, Z., Huang, Z., Li, Z., Zhu, X., Hu, X., Lu, T., Lu, L., Li, H., et~al.: Internimage: Exploring large-scale vision foundation models with deformable convolutions. In: Proceedings of the IEEE/CVF Conference on Computer Vision and Pattern Recognition. pp. 14408--14419 (2023)

\bibitem{wang2020deep}
Wang, X., Ma, H., You, S.: Deep clustering for weakly-supervised semantic segmentation in autonomous driving scenes. Neurocomputing  \textbf{381},  20--28 (2020)

\bibitem{wang2023images}
Wang, X., Wang, W., Cao, Y., Shen, C., Huang, T.: Images speak in images: A generalist painter for in-context visual learning. In: Proceedings of the IEEE/CVF Conference on Computer Vision and Pattern Recognition. pp. 6830--6839 (2023)

\bibitem{wang2023SegGPT}
Wang, X., Zhang, X., Cao, Y., Wang, W., Shen, C., Huang, T.: Seggpt: Segmenting everything in context. arXiv preprint arXiv:2304.03284  (2023)

\bibitem{wang2022unsupervised}
Wang, X., Lian, L., Yu, S.X.: Unsupervised selective labeling for more effective semi-supervised learning. In: European Conference on Computer Vision. pp. 427--445. Springer (2022)

\bibitem{wang2004image}
Wang, Z., Bovik, A.C., Sheikh, H.R., Simoncelli, E.P.: Image quality assessment: from error visibility to structural similarity. IEEE transactions on image processing  \textbf{13}(4),  600--612 (2004)

\bibitem{waqas2019isaid}
Waqas~Zamir, S., Arora, A., Gupta, A., Khan, S., Sun, G., Shahbaz~Khan, F., Zhu, F., Shao, L., Xia, G.S., Bai, X.: isaid: A large-scale dataset for instance segmentation in aerial images. In: Proceedings of the IEEE/CVF Conference on Computer Vision and Pattern Recognition Workshops. pp. 28--37 (2019)

\bibitem{wei2022chain}
Wei, J., Wang, X., Schuurmans, D., Bosma, M., Xia, F., Chi, E., Le, Q.V., Zhou, D., et~al.: Chain-of-thought prompting elicits reasoning in large language models. Advances in Neural Information Processing Systems  \textbf{35},  24824--24837 (2022)

\bibitem{yao2021scale}
Yao, X., Cao, Q., Feng, X., Cheng, G., Han, J.: Scale-aware detailed matching for few-shot aerial image semantic segmentation. IEEE Transactions on Geoscience and Remote Sensing  \textbf{60},  1--11 (2021)

\bibitem{ye2023compositional}
Ye, J., Wu, Z., Feng, J., Yu, T., Kong, L.: Compositional exemplars for in-context learning. arXiv preprint arXiv:2302.05698  (2023)

\bibitem{yu2021crossover}
Yu, Q., Gao, Y., Zheng, Y., Zhu, J., Dai, Y., Shi, Y.: Crossover-net: Leveraging vertical-horizontal crossover relation for robust medical image segmentation. Pattern Recognition  \textbf{113},  107756 (2021)

\bibitem{yu2023h2rbox}
Yu, Y., Yang, X., Li, Q., Zhou, Y., Da, F., Yan, J.: H2rbox-v2: Incorporating symmetry for boosting horizontal box supervised oriented object detection. In: Thirty-seventh Conference on Neural Information Processing Systems (2023)

\bibitem{zendel2022unifying}
Zendel, O., Sch{\"o}rghuber, M., Rainer, B., Murschitz, M., Beleznai, C.: Unifying panoptic segmentation for autonomous driving. In: Proceedings of the IEEE/CVF Conference on Computer Vision and Pattern Recognition. pp. 21351--21360 (2022)

\bibitem{zhang2011fsim}
Zhang, L., Zhang, L., Mou, X., Zhang, D.: Fsim: A feature similarity index for image quality assessment. IEEE transactions on Image Processing  \textbf{20}(8),  2378--2386 (2011)

\bibitem{zhang2023makes}
Zhang, Y., Zhou, K., Liu, Z.: What makes good examples for visual in-context learning? arXiv preprint arXiv:2301.13670  (2023)

\bibitem{zhang2021transfuse}
Zhang, Y., Liu, H., Hu, Q.: Transfuse: Fusing transformers and cnns for medical image segmentation. In: Medical Image Computing and Computer Assisted Intervention--MICCAI 2021: 24th International Conference, Strasbourg, France, September 27--October 1, 2021, Proceedings, Part I 24. pp. 14--24. Springer (2021)

\bibitem{zhang2022automatic}
Zhang, Z., Zhang, A., Li, M., Smola, A.: Automatic chain of thought prompting in large language models. arXiv preprint arXiv:2210.03493  (2022)

\end{thebibliography}

\clearpage

\end{document}